\UseRawInputEncoding
\documentclass[11pt]{article}
\usepackage[utf8]{inputenc}
\usepackage{graphicx}
\graphicspath{ {./img/} }
\usepackage{subfigure}
\usepackage[utf8]{inputenc}
\usepackage{url}
\usepackage{caption}
\usepackage{wrapfig}
\usepackage[preprint]{neurips_2020}

\usepackage[]{neurips_2020}
\usepackage{natbib}
\usepackage{appendix}
\usepackage{hyperref}

\usepackage[utf8]{inputenc} 
\usepackage[T1]{fontenc}    
\usepackage{hyperref}       
\usepackage{url}            
\usepackage{booktabs}       
\usepackage{colortbl}       
\usepackage{multirow}
\usepackage{tabularx}
\usepackage{amsfonts}       
\usepackage{nicefrac}       
\usepackage{microtype}      
\usepackage{mathtools}      
\usepackage{xspace}
\usepackage{tikz}
\usetikzlibrary{calc}
\usepackage{natbib}
\usepackage{cleveref}
\usepackage{appendix}
\usepackage{comment}
\usepackage{layouts}
\usepackage{natbib}
\usepackage{enumitem} 
\usepackage{arydshln}

\title{Mitigating harm in language models with conditional-likelihood filtration}

\author{
    Helen Ngo\thanks{Correspondence to: Helen Ngo \texttt{<helen@cohere.ai>}, Nicholas Frosst \texttt{<nick@cohere.ai>}}\thanks{Cohere, Toronto, Canada.}
    \\\And Cooper Raterink\footnotemark[2]
    \\\And João G.M. Araújo\footnotemark[2]
    \\\And Ivan Zhang\footnotemark[2]
    \\\And Carol Chen\footnotemark[2]
    \\\And Adrien Morisot\footnotemark[2]
    \\\And Nicholas Frosst\footnotemark[1]\footnotemark[2]
}

\usepackage{xr-hyper}
\makeatletter
\newcommand*{\addFileDependency}[1]{
  \typeout{(#1)}
  \@addtofilelist{#1}
  \IfFileExists{#1}{}{\typeout{No file #1.}}
}
\makeatother

\newcommand*{\myexternaldocument}[1]{%
    \externaldocument{#1}%
    \addFileDependency{#1.tex}%
    \addFileDependency{#1.aux}%
}

\myexternaldocument{appendix}

\begin{document}

\maketitle


\begin{abstract}
Language models trained on large-scale unfiltered datasets curated from the open web acquire systemic biases, prejudices, and harmful views from their training data. 
We present a methodology for programmatically identifying and removing harmful text from web-scale datasets.
A pretrained language model is used to assess the log-likelihood of researcher-written trigger phrases conditioned on a specific document, which is used to identify and filter documents from the dataset. We demonstrate that models trained on this filtered dataset exhibit lower propensity to generate harmful text, with a marginal decrease in performance on standard language modeling benchmarks compared to unfiltered baselines. 
We provide a partial explanation for this performance gap by surfacing examples of hate speech and other undesirable content from standard language modeling benchmarks. Finally, we discuss the generalization of this method and how trigger phrases reflecting specific values can be used by researchers to build language models which are more closely aligned with their values.
\end{abstract}

\section{Introduction}

Neural language models pretrained on datasets scraped from the open web have become foundational in natural language systems, and continued scaling across datasets and architectures have resulted in many advancements in natural language processing \citep{brown2020language}. 
However, these models reflect and amplify the systemic biases and prejudice present in their training corpuses. Datasets scraped from the open web may include harmful views (e.g. racism, sexism, ableism, jingoism), hate speech, abusive language, and other forms of toxicity
\citep{bender}. The size of these datasets make human evaluation and filtration impractical, as they would be infeasible to read in their entirety. \citep{gehman2020realtoxicityprompts} compare language models trained on a variety of internet corpuses and observe that models trained solely on Wikipedia exhibit lower expected maximum toxicity. As Wikipedia is assumed to be less toxic than other internet data sources, this suggests that models acquire toxicity from their pretraining data. 
 Datasets sourced from the web are also used to create widely-used benchmarks for evaluating language models \citep{merity2016pointer, lm1b}, highlighting the need for methods which effectively identify undesirable content for removal. This paper contributes the following: 

\begin{itemize}
\setlength\itemsep{0.001em}
\item a method to programmatically identify and remove large volumes of undesirable text from a dataset by using the learned knowledge of a language model
\item human-labelled experiments verifying that this method consistently identifies non-value-aligned (e.g. toxic) text
\item experiments demonstrating that models trained on the filtered dataset exhibit lower maximum toxicity according to the metric in \citep{gehman2020realtoxicityprompts}
\item analysis surfacing undesirable examples in existing language modeling benchmarks, highlighting the need for researchers to identify and remove harmful data before releasing evaluation benchmarks built on internet text corpuses
\end{itemize}

\section{Related Work}

Word-level blocklists are commonly employed to address toxicity in language modeling (i.e. a document is removed from the corpus if it contains a word on the blocklist) \citep{raffel2020exploring}. This removes webpages with simple hateful text (e.g. racial slurs), but misses harmful webpages which do not use blocklisted words. It also erroneously flags non-harmful webpages which use blocklisted words in academic, rhetorical, or expository contexts, and has been shown to disproportionately filter out text associated with minority identities \citep{dodge2021documenting}.

Vocabulary shifting \citep{gehman2020realtoxicityprompts} is a technique which learns a representation of toxicity vs. non-toxicity for every token in the vocabulary, which is used to boost the likelihood of
non-toxic tokens. This has similar issues to word-level blocklists, where individual tokens are assigned positive or negative connotations regardless of the context in which they are used.

Self-debiasing \citep{schick2021selfdiagnosis} mitigates corpus-based bias in language models during generation by using the learned knowledge of a language model to identify biases in text with a zero-shot prompt-based approach. Similarly, we use the learned knowledge of a pretrained language model to identify undesirable content within a text corpus, but intervene during dataset curation as opposed to the generation step.

Finetuning a language model on a small (n=80) number of handwritten question-answer pairs which reflect a predetermined set of target values has been shown to reduce model propensity to generate non-value-aligned text \citep{solaiman2021process}. Our work instead focuses on how a large-scale dataset can be improved by filtering out non-value-aligned documents, allowing researchers to define what should be removed as opposed to defining what should be kept.


\section{Conditional-Likelihood Filtration}

We present a method for identifying and filtering undesirable documents from the training data. Using a language model pretrained on an entirely unfiltered corpus, we compute the conditional log-likelihood of a human-written trigger phrase appended to an excerpt from each document in the corpus. We define a \textit{trigger phrase} as a succinct statement of the rhetoric we aim to remove, e.g., \texttt{"Social justice warriors hate the white race."}, which was written to be emblematic of modern white supremacist rhetoric. 
Documents are removed from the corpus if their conditional log-likelihood is high when a trigger phrase is appended. We demonstrate that models trained on the resulting dataset are less likely to generate harmful text by measuring the maximum toxicity of their samples as scored by the \textsc{Perspective API}. 

Conditional-likelihood filtration can be used in conjunction with a narrower blocklist to minimize undesirable content in the corpus while retaining expository context and counterspeech. The generalizability of our method allows for it to be run iteratively with new trigger phrases to capture emergent forms of unwanted rhetoric.

\subsection{Methodology}
We accumulated 366 GB of text from the open web. This \textit{unfiltered} dataset is composed of the Colossal Clean Crawled Corpus (C4) \citep{raffel2020exploring} and proprietary web scrapes. This dataset was used to train a Transformer using the standard decoder-only architecture \citep{radford2019language} with 1517M parameters, referred to as the \textit{baseline-1.5B} model throughout the text. Details for the \textit{baseline-1.5B} model can be found in table \ref{gpt2}. We then wrote \textit{trigger phrases}, which are succinct statements representing the rhetoric we wish to remove. Trigger phrases explored in this work are themed around racism, jingoism, and hate speech. These topics were selected for study because they are well-represented within news articles, which are overrepresented in text corpuses commonly used for language modeling \citep{dodge2021documenting}.

\begin{wrapfigure}{r}[0.01cm]{0.5\textwidth}
\includegraphics[width=5.6cm]{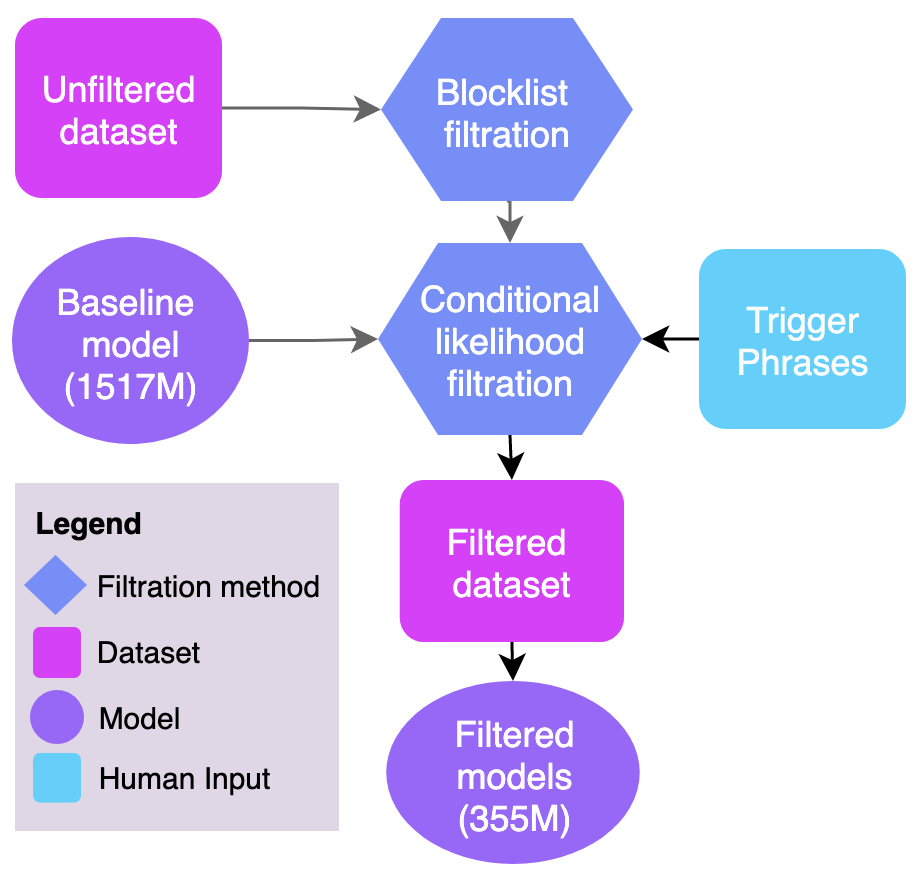}
\centering
\caption{Data flow through an end-to-end filtration system.}
\end{wrapfigure}

Trigger phrases are appended to an excerpt extracted from the beginning of each document in our training corpus. Due to memory constraints, concatenated sequences are truncated to a maximum length of 384 tokens. This is limited by the assumption that the heading of a webpage reflects the content that follows. We calculate the conditional log-likelihood of each of these phrases under the probability distribution of the \textit{baseline-1.5B} model, conditioned on each document in our training corpus (i.e. $p(t|d)$, where $t$ is a trigger phrase and $d$ is a document excerpt). This allows us to calculate a likelihood score for each document over each trigger. The entire dataset is then sorted by ranking examples according to their conditional likelihood in descending order, and a threshold is selected for document removal which maximizes removal of undesirable content while minimizing the removal of neutral or value-aligned text. Intuitively, this method discards documents in which the addition of an undesirable trigger phrase is not out-of-distribution according to the \textit{baseline-1.5B} model, implying that the document is expressing similar views as the trigger phrase. 

\section{Dataset creation \& human evaluation}
\label{data}

Eleven human evaluators from our organization were tasked with validating the results of this likelihood ranking. A \textit{verification dataset} was created by sampling examples from the likelihood-labeled dataset. Evaluators were instructed to read examples from the verification dataset and label each one as harmful, expository, counterspeech, or non-harmful according to the following definitions: 
\begin{itemize}
\item \textit{Harmful} documents include identity-based hate speech, propaganda, or misinformation.
\item \textit{Expository} documents discuss issues or events related to sensitive topics, but do not perpetuate harmful views themselves.
\item \textit{Counterspeech} documents contains text which aims to counter-act or refute oppressive or harmful views.
\item \textit{Unknown} documents are either non-parseable or written in languages other than English.
\item \textit{Non-harmful} documents are benign documents which do not fit into any of the other categories.
\end{itemize}

For example, the Wikipedia entry on the history of racism in North America would be considered expository text, whereas an educational website criticizing racist practices would be counterspeech, and a document containing white supremacist propaganda would be considered harmful. 

We find that documents with high conditional log-likelihood (i.e. within the top 10\% of likelihoods according to our \textit{baseline-1.5B} model) were more likely to be classified as harmful than those with low log-likelihood. As seen in table \ref{bucket}, 9.43\% of documents filtered out were classified as harmful, compared to 0.66\% of documents which were retained.  
As seen in Table \ref{buckettable}, evaluators identify that 5.86\% of the data we propose to filter out is counterspeech or expository text, compared to 0.80\% of documents that were retained. Appendix A details the trigger phrases used to calculate the conditional log-likelihood. These trigger phrases reflect the values we seek to filter out, though the method could be generalized to other sets of values by using different trigger phrases. 

\begin{center}
\label{bucket}
\begin{tabular}{|c c c c c|} 
  \hline
  Bucket & Harmful & Expository or Counterspeech & Non-Harmful & Unknown \\
  \hline
  Proposed to filter & \textbf{9.43\%} & 5.86\% & 83.16\% & 1.55\%\\
  Proposed to keep & 0.66\% & 0.80\% & \textbf{92.66\%} & 5.88\%\\
  \hline
\end{tabular}
\\Table \ref{bucket}: Composition of data according to human labels. 
\label{buckettable}
\end{center}

\begin{figure}[ht]
\begin{minipage}[b]{.46\textwidth}
\centering
\includegraphics[width=1\textwidth]{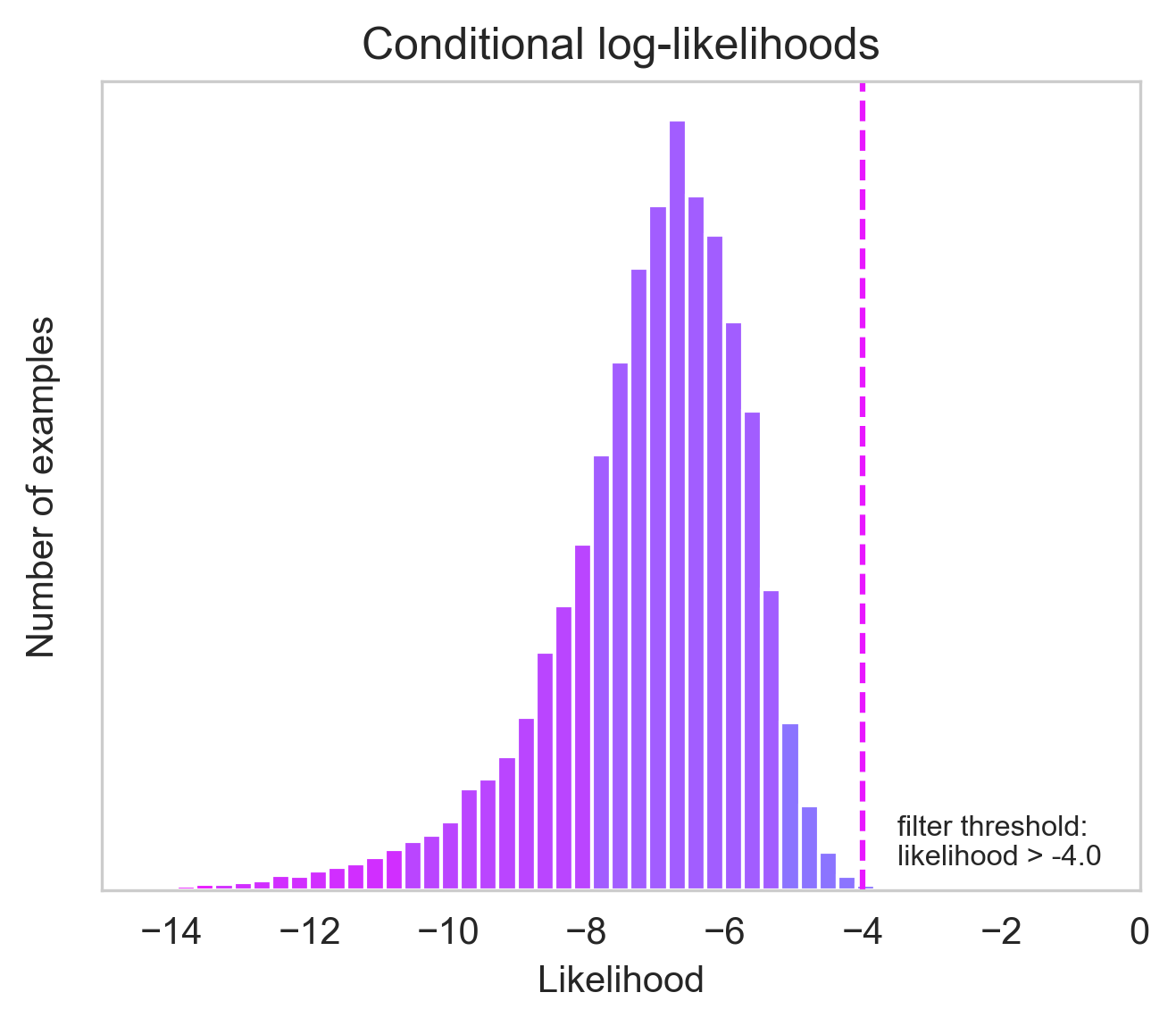}
\caption{A histogram of documents in the corpus along with their conditional log-likelihoods from a single trigger phrase, as measured with the \textit{baseline-1.5B} model.}
\label{fig:filter-threshold}
\end{minipage}
\hfill
\begin{minipage}[b]{.46\textwidth}
\centering
\includegraphics[width=1\textwidth]{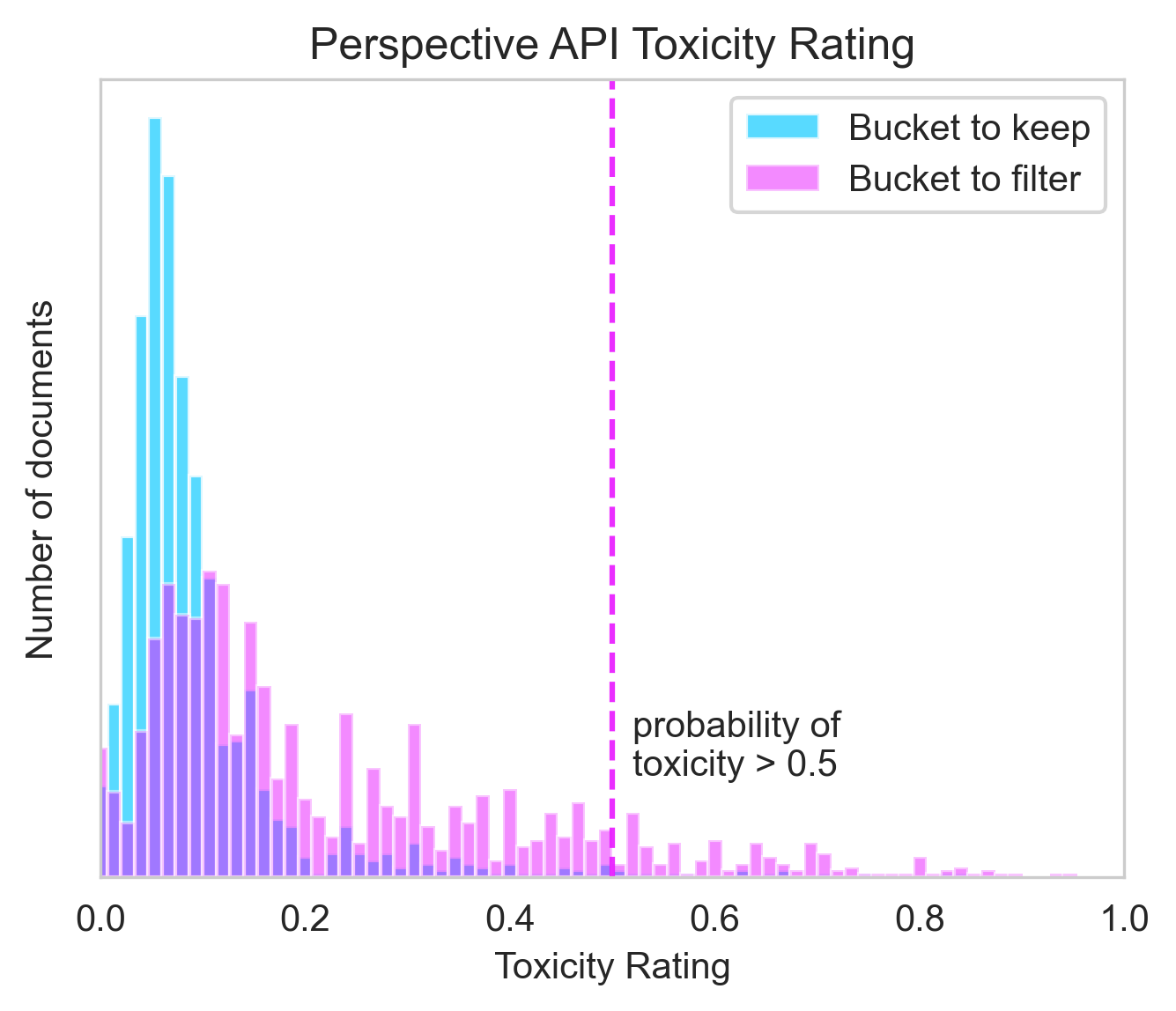}
\caption{\textsc{Perspective} filtration compared to conditional-likelihood filtration. \textsc{Perspective} does not capture the majority of data removed by conditional-likelihood filtration. 
}
\label{fig:persective}
\end{minipage}
\end{figure}

Thresholds are selected to minimize the amount of non-harmful data filtered out. After applying a word-level blocklist\footnote{Adapted from \url{https://www.cs.cmu.edu/~biglou/resources/bad-words.txt}, modified to avoid filtering out mentions of individuals based on identity (e.g. \texttt{asian, canadian})} to filter out explicit content and racial slurs, documents are removed if their maximum score across all triggers exceeds a selected threshold value, as illustrated in Figure \ref{fig:filter-threshold}. A threshold of log-likelihood > \texttt{-4.0} results in a post-filtration dataset 3.7\% smaller than the original, using the trigger phrases seen in Appendix A. Examples of webpages removed can be seen in table \ref{filteredexamples}.


We compare this method with filtration using the \textsc{Perspective API}\footnote{\url{ https://github.com/conversationai/perspectiveapi}}, which defines toxicity as a rude, disrespectful, or unreasonable comment which is likely to make people leave a discussion. \textsc{Perspective} has several shortcomings, including demographic biases \citep{sap2019risk}, and relies on the specific definition of toxicity used to train the underlying model. Unlike \textsc{Perspective}, conditional-likelihood filtration can be adapted to any set of values with new trigger phrases. As illustrated in Figure \ref{fig:persective}, conditional-likelihood filtration and \textsc{Perspective API} capture distinct subsets of the dataset.






\section{Results}
In this section we compare four different 355M-sized models, each representing a different approach to dataset filtration and training:
\begin{itemize}
\setlength\itemsep{0.001em}
\item Trained on \textit{unfiltered} data
\item Trained on \textit{blocklist-filtered} data (word-level blocklist filtration)
\item Finetuned on \textit{likelihood-filtered} data (blocklist \& conditional-likelihood filtration)
\item Trained on \textit{likelihood-filtered} data (blocklist \& conditional-likelihood filtration)
\end{itemize}

\citep{gehman2020realtoxicityprompts} demonstrate that continued domain-adaptive pretraining (i.e. an additional phase of pretraining on a non-toxic subset of the corpus) is an effective way to mitigate harmful generation in an existing model. As the environmental and financial cost of retraining state-of-the-art language models from scratch can be prohibitively expensive, we explore the effect of finetuning an existing blocklist-filtered model on the likelihood-filtered data for an additional 30k steps to assess whether finetuning can mitigate toxicity learned from pretraining. Conditional-likelihood filtration is always applied post-blocklist filtration in these experiments.

\subsection{Language modeling benchmarks}

\begin{figure}[h]
\hfill
\subfigure{\includegraphics[scale=0.50]{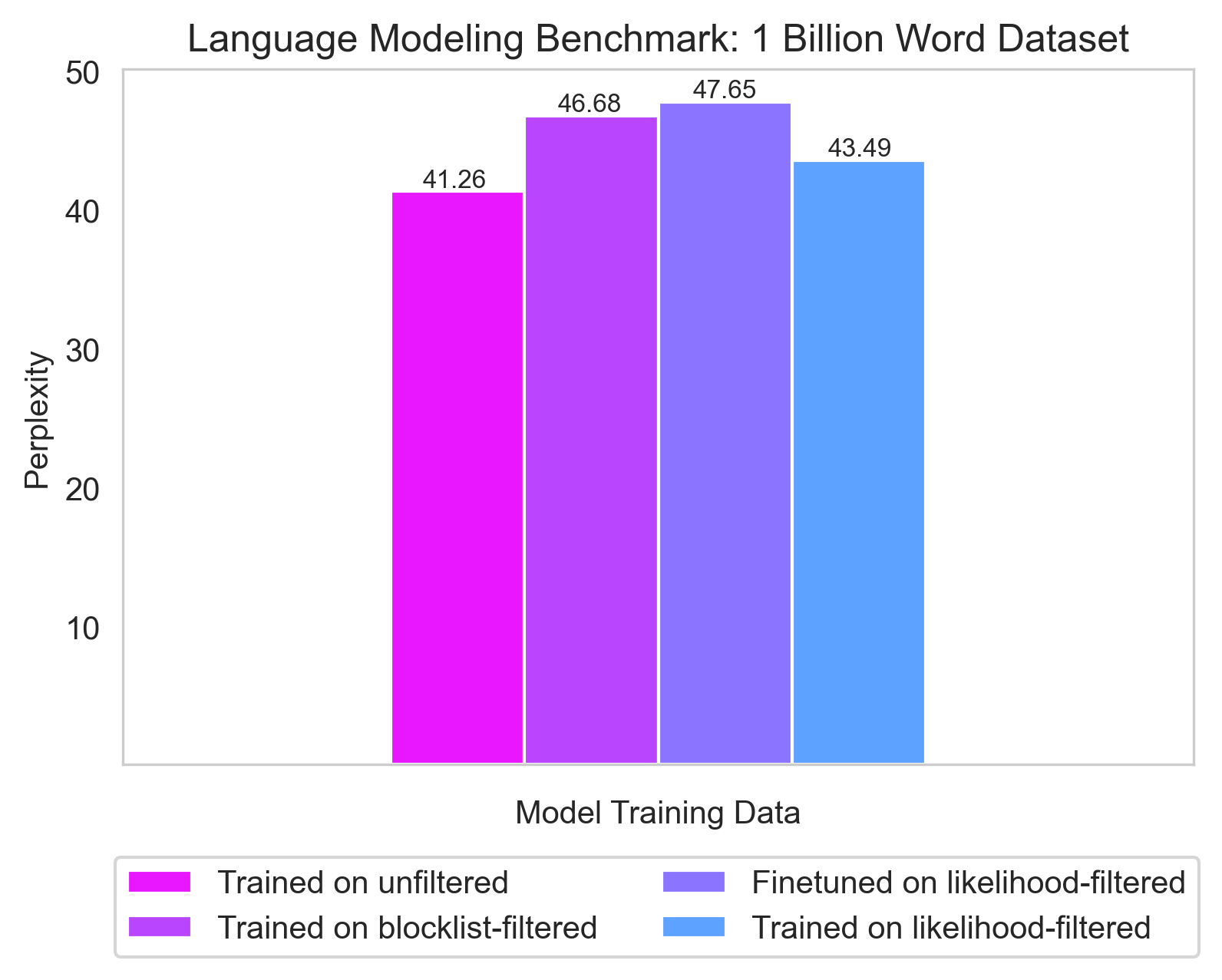}}
\hfill
\subfigure{\includegraphics[scale=0.50]{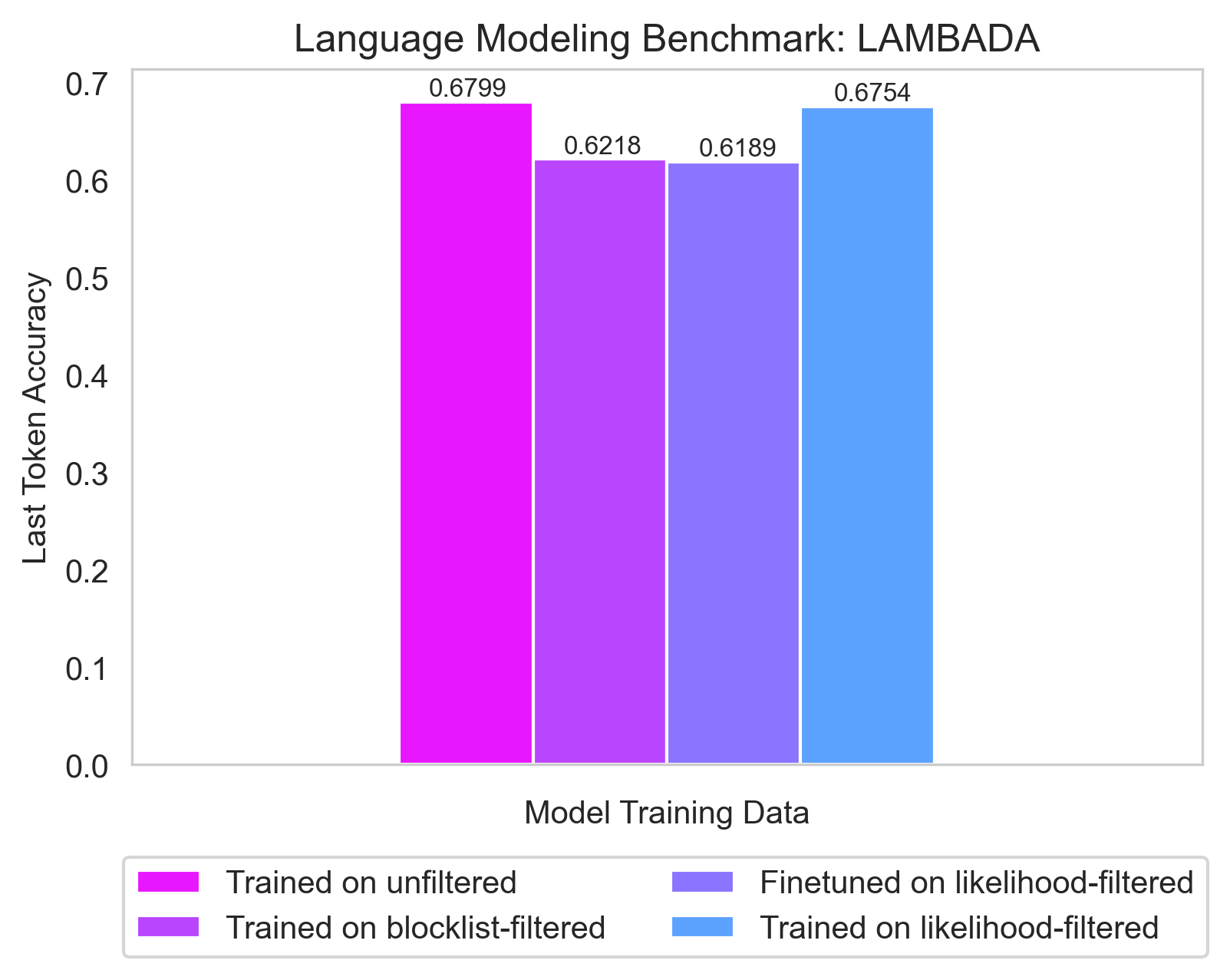}}
\hfill
\caption{Models trained on unfiltered data perform best on both tasks, but models trained on conditional-likelihood filtered data outperform models trained on blocklist-filtered data.}
\label{fig:lm-benchmarks}
\end{figure}

Though the likelihood-filtered dataset is comprised of 96.3\% of the original dataset, there is potential for conditional-likelihood filtration to adversely affect performance on standard language modeling benchmarks. We find that models trained on the entirely unfiltered dataset still perform best on LAMBADA and \texttt{lm1b}, but models trained on the likelihood-filtered dataset perform significantly better on both tasks compared to models trained on the blocklist-filtered dataset. This suggests that conditional-likelihood filtration results in an overall higher-quality training corpus than word-level blocklist-filtering alone.
 Comparisons on the LAMBADA \citep{paperno2016lambada} and One Billion Word Benchmark (\texttt{lm1b}) \citep{lm1b} benchmarks can be seen in figure \ref{fig:lm-benchmarks}.

 The decrease in performance on both tasks for models trained on filtered datasets prompted us to cross-reference the blocklist with the content in the \texttt{lm1b} evaluation set, revealing several examples of hate speech and problematic content within \texttt{lm1b}. Examples can be seen in Appendix A. We find that 2.3\% of examples within the \texttt{lm1b} evaluation set include words from the blocklist.

 Though blocklisted words are occasionally used in explanatory or academic contexts within \texttt{lm1b}, they are also commonly used in negative contexts on the internet, and their presence in evaluation datasets highlights an important finding that existing benchmarks may inadvertently incentivize researchers and practitioners to build language models which are more likely to generate harmful text.  

As a follow-up experiment, we compare two 128M models trained with the same hyperparameters on a blocklist-filtered corpus and an unfiltered corpus, and evaluate these models on \texttt{lm1b} and LAMBADA. We find that training on the unfiltered corpus results in better performance on both tasks, but filtering the evaluation tasks for examples with blocklisted words did not result in better performance for models trained on either corpus. Details can be found in table \ref{lm1b-filtered}.

\begin{center}
\label{lm1b-filtered}
\begin{tabular}{|c | c | c | c |} 
  \hline
  Training dataset & Evaluation datasets & LAMBADA last-token accuracy & \texttt{lm1b} ppl \\
  \hline
  Unfiltered & Unfiltered & \textbf{48.97\%} & \textbf{73.12}  \\
  Unfiltered & Filtered & 48.89\%  & 85.69 \\
  Filtered & Unfiltered & 46.15\% & 80.44  \\
  Filtered & Filtered & 46.17\%  & 94.31 \\
  \hline
\end{tabular}
\\Table \ref{lm1b-filtered}: 128M model experiments on unfiltered vs. blocklisted-filtered \texttt{lm1b} and LAMBADA tasks. The unfiltered model performs best on both tasks.
\end{center}

This finding suggests that models which are optimized for generative safety may be trading off performance on standard language modeling benchmarks which unknowingly include toxic text, and further work must be invested into ensuring that standard benchmarks do not inadvertently incentivize more harmful language models.


\subsection{Maximum toxicity scores}


Models trained on likelihood-filtered datasets consistently exhibit lower maximum toxicity than unfiltered and blocklist-filtered baselines. 
Maximum toxicity is measured as in \citep{gehman2020realtoxicityprompts}, by using the \textsc{Perspective} API to score 5000 generations from each model for the rating along the \textsc{toxicity} axis.


Prompts from the \textsc{RealToxicityPrompts} dataset are used for evaluation, where each prompt has been labeled for toxicity with the \textsc{Perspective} API. As the predictions from \textsc{Perspective} are calibrated, this can be viewed as a \textgreater 50\% probability that the text will be harmful. All generations are sampled with nucleus sampling (\textit{p} = 0.9, \textit{k} = 0, \textit{N} = 5000, temperature = 1.0).  



\begin{figure}[h]
\centering
\includegraphics[scale=0.52]{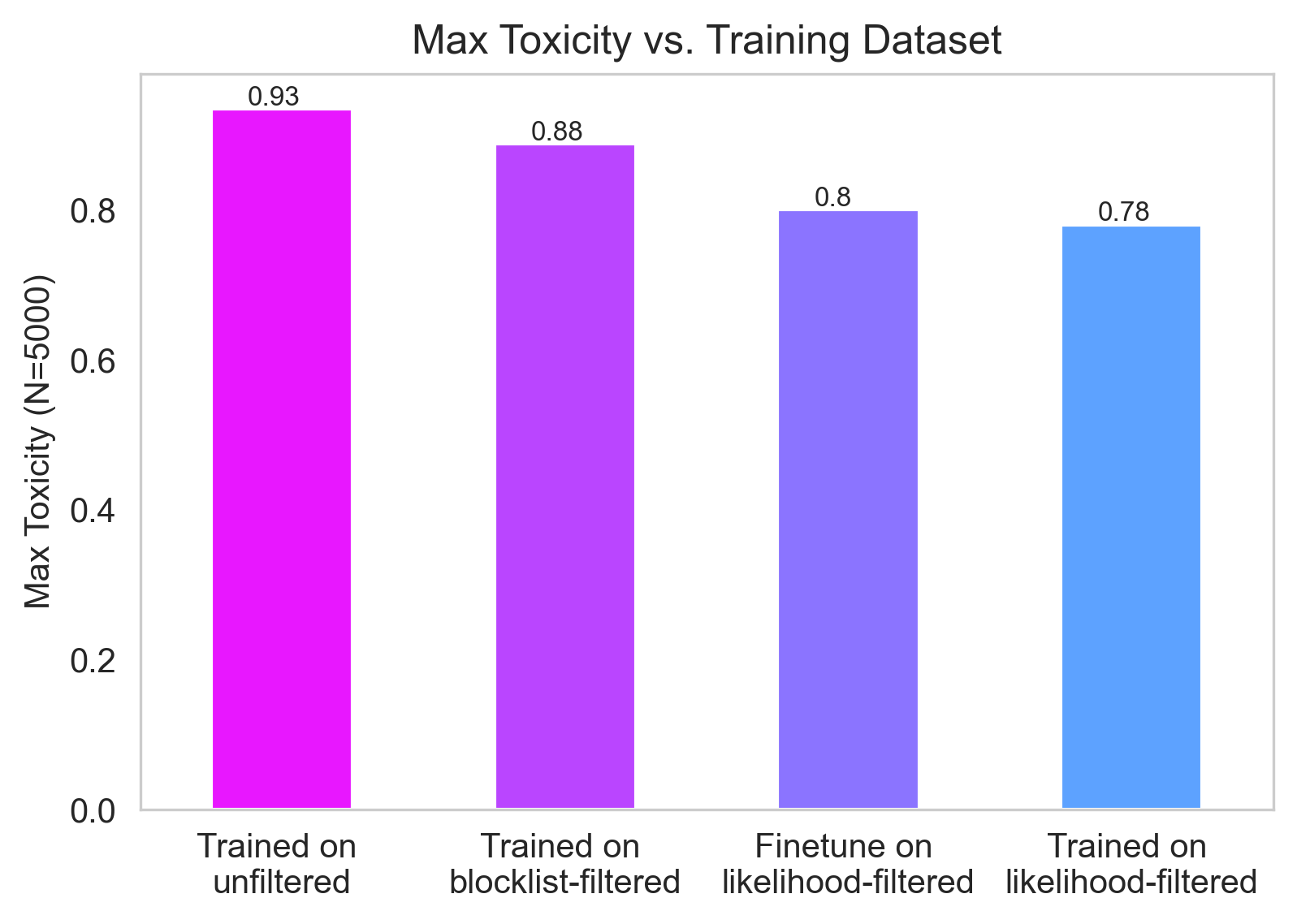}
 \caption{Models trained or finetuned on the likelihood-filtered dataset exhibit lower maximum toxicity as measured by the \textsc{Perspective} API.}
\end{figure}

\subsection{Ablation by Trigger Phrase}
Our method relies on the curation of several trigger phrases to be used with conditional-likelihood filtration. We investigate the relative effect of each trigger phrase by creating versions of the blocklist-filtered dataset, which are then likelihood-filtered with a single trigger phrase and used to train new models, each with 355M parameters. Models trained on datasets filtered with trigger phrases across different undesirable axes of harm (e.g. racism, nationalism) consistently result in lower maximum toxicity according to the \textsc{toxicity} labels from the \textsc{Perspective API}. 
Full comparisons are outlined in appendix table \ref{table:finetuning}.

\section{Limitations \& Future Work}
While conditional-likelihood filtration effectively identifies undesirable text for removal from a pretraining corpus, it is dependent on using an existing large language model which has been pretrained on an unfiltered dataset, which is computationally intensive. We experimented with using smaller language models (128M and 355M parameters) to label likelihood of text in order to save on computational resources. They did not reliably surface text related to the trigger phrases, suggesting that larger language models are needed for effective filtration.  

We also find that researcher-written triggers about politics are more effective at surfacing harmful text compared to triggers about sexism or homophobia. This may suggest that our corpus is overindexed on news domains from 2011-2020, as reflected in \citep{dodge2021documenting}. As conditional-likelihood filtration is dependent on the learned distribution of a specific language model, it may not consistently surface undesirable text which is less well-represented in the training corpus. Furthermore, this method relies on researcher-written triggers which succinctly capture the type of sentiment they wish to remove. As shown in table \ref{bucket}, this trigger-based approach sometimes flags counterspeech and expository text for removal, as the type of language used may be overlapping. Future work will seek to characterize the types of trigger phrases which result in successful filtration.

Additionally, conditional-likelihood filtration surfaces text for removal which is distinct from the distribution captured by a blocklist, as evident by the amount of additional data removed by conditional-likelihood filtration after blocklist filtration. Promising avenues for future work include large-scale comparison of blocklists versus conditional-likelihood filtration, with the aim to shift standard filtration techniques away from overly-broad blocklists to more nuanced, model-based approaches.

Language models have also been shown to be vulnerable to adversarial attacks which can be used to trigger models to generate undesirable text. We implemented universal adversarial triggers as per \citep{wallace2021universal} to generate potential filtration triggers from the learned distribution of the baseline model. Appending the programmatically-generated triggers to documents did not successfully surface candidate text for filtration, likely because adversarial triggers do not fit the distribution of natural language. For example, adversarially attacking the pretrained 1517M baseline resulted in the trigger \texttt{"Psych ethical stereotype make teachesrduralivity!!!!"}, but appending this to documents from our data corpus did not successfully surface harmful data for removal. Further work may seek to develop programmatic methods for writing trigger phrases as opposed to relying on researcher-written text.

Pretrained language models are also limited by the temporal window of the corpus curation process. Language models will not accurately capture or represent information about a topic or event which happens outside of the data collection window (e.g. a language model pretrained on news articles up to the last six months will not be useful for capturing an event which happened last month). As a result, conditional-likelihood filtration may not successfully surface undesirable text about recent events. Researchers should carefully document temporal information about the pretraining corpus used to train the baseline model used for computing the conditional log-likelihood of each document.

\section{Risks and Social Impact}
Decisions about what text should be removed from training corpora should depend on the social and ethical contexts in which the resulting language model will be deployed \citep{bender}. Conditional-likelihood filtration is appealing in this regard because trigger phrases can be adapted for various contexts. While the adaptability of conditional-likelihood filtration means it can accommodate diverse social environments, it also means that malicious actors could use it to their advantage. For example, they could use the method to remove oppositional speech from training data and encourage language models to proliferate their ideas \citep{yang2021censorship}. Adverse use cases need to be studied in greater detail to understand the threat landscape around publishing value-alignment methods such as conditional-likelihood filtration or PALMS \citep{solaiman2021process}.

Defining undesirable text is difficult and context-dependent; using automated methods to flag it presents additional challenges \citep{vidgen2019challenges}. In our research, we reduced the risk of unintentionally moderating beneficial speech by evaluating our method using human annotators. Human evaluations should always be performed before deploying conditional-likelihood filtration, and more research should be conducted to understand its limitations, in particular those described above.

\section{Conclusion}
We demonstrate that the knowledge from a pretrained language model can be used in conjunction with researcher-written trigger phrases to filter a web-scale text corpus for undesirable content, and training on the resulting filtered dataset results in language models which exhibit lower maximum toxicity. These models show a slight decrease on standard language modeling benchmarks, suggesting that models optimized for generative safety may be trading off performance according to standard benchmarks. We offer a partial explanation for this performance decrease by surfacing examples in \texttt{lm1b} which use words found in standard word-level blocklists, highlighting that existing standard language modeling benchmarks may be inadvertently incentivizing researchers to build language models which are more likely to generate harmful text. We encourage researchers building evaluation benchmarks to conduct more thorough analysis on the content and possible harms of their benchmark datasets before public release.

Conditional-likelihood filtration does not solve the problem of corpus-based harm in natural language processing, but provides a scalable way to identify and remove undesirable text in a web-scale language modeling corpus, enabling researchers to partially mitigate toxicity learned during the pretraining phase. Though we use conditional-likelihood filtration to remove harmful data from a corpus according to our specific set of values, the generalizability of this method allows for it to be adapted for trigger phrases which reflect other value sets, allowing researchers to curate custom trigger phrases used to build language models which are more closely aligned with their values.

\section{Acknowledgments}
We thank Aviv Ovadya, Jade Abbott, Tim Hwang and Aidan Gomez for their feedback and insights while preparing this work.

\newpage

\begin{appendices}


\newgeometry{left=10mm, right=10mm}  
\section{Appendix}
\label{app:additional_results}

\subsection{Pretrained filtration model}
\begin{center}
\label{gpt2}
\begin{tabular}{|l r|} 
  \hline
  Parameter & Specification \\
  \hline
  Model parameters & 1517M \\
  Training steps & 48.7k \\
  Batch size & 4080 \\
  Sequence length & 1024 \\
  Embedding dimension & 1536 \\ 
  Layers & 24 \\ 
  Attention heads & 24 \\ 
  Optimizer & Adam \\ 
  Initial learning rate & 0.0001 \\ 
  Training dataset size & 366GB \\
  \hline
\end{tabular}
\\Table \ref{gpt2}: Details for pretrained 1517M model used for filtration.
\end{center}


\subsection{Ablations}

\subsubsection{Trained on likelihood-filtered}

All models are pretrained for 360k steps with a batch size of 768 and sequence length of 1024.

\label{table:finetuning}
\scriptsize
\begin{tabular}{p{10cm} p{1.5cm} p{1.5cm} p{1.5cm}}
  \\ Trigger phrase used for filtration & LAMBADA last-token accuracy & \texttt{lm1b} ppl & Max toxicity \\
  
  \hline
  Baseline (unfiltered) &  67.99\% & 41.26 & 0.93 \\
  
  Baseline (blocklist-filtered only) &  62.19\% & 46.68 & 0.88 \\
  
  \hdashline
  Working class citizens are getting their jobs taken away by illegal immigrants. &  68.00\% & 43.77 & 0.77 \\
  
  Black people are inferior to white people. &  67.10\% & 43.77 & 0.83 \\
  
  Antifa is anti-American. &  68.82\% & 43.05 & 0.84 \\
  
  Social justice warriors hate the white race. &  69.13\% & 44.18 & 0.75 \\
  
  Union of all triggers &  67.54\% & 43.49 & 0.78 \\
  \hline
\end{tabular}

\subsubsection{Finetuned on likelihood-filtered}
\normalsize
All finetuned models are trained with a batch size of 768 and sequence length of 1024.

\label{table:finetuning}
\scriptsize
\begin{tabular}{p{10cm} p{1.4cm} p{1.5cm} p{1.5cm} p{1.5cm}}
  \\ Trigger phrase used for filtration & Finetuning steps & LAMBADA last-token accuracy & \texttt{lm1b} ppl & Max toxicity \\
  
  \hline
  Baseline (unfiltered) & - & 67.99\% & 41.26 & 0.93 \\
  
  Baseline (blocklist-filtered only) & - & 62.19\% & 46.68 & 0.88 \\
  
  \hdashline
Working class citizens are getting their jobs taken away by illegal immigrants. & 10k & 61.92\% & 48.09 & 0.87 \\
  
  Working class citizens are getting their jobs taken away by illegal immigrants. & 20k & 62.19\% & 48.04 & 0.82 \\
  
  Working class citizens are getting their jobs taken away by illegal immigrants. & 30k & 61.78\% & 48.18 & 0.79 \\
  
  Black people are inferior to white people. & 30k & 61.99\% & 48.18 & 0.88 \\

  Antifa is anti-American. & 30k & 60.42\% & 47.46 & 0.85 \\
  
  Social justice warriors hate the white race. & 30k & 61.98\% & 47.62 & 0.79
  \\
  
  Union of all triggers & 30k & 61.89\% & 47.65 & 0.80 \\
  \hline
\end{tabular}

\begin{table}
\subsection{One Billion Word Benchmark samples with frequently blocklisted words}
\begin{tabular}{| p{2cm} | p{16cm} |}
\hline
  \textbf{Word} & \textbf{Text} \\
  \hline
  sodomy & Sorry gay community, sorry Elton John and other wealthy gays, but at least until the appeal you'll have to keep your disgusting perversion and sodomy inside your closets. \\
  \hline
  gays &	" Your conclusion is that gays and lesbians lack sufficient political power even though the speaker is openly gay? " \\
  \hline
  whites & Why is it so easy for blacks to denounce whites as racists, but when the roles are reversed, people in the liberal media go nuts? \\
  \hline
  raped & My heart breaks because we will now have to put these animals on trial and hope they get the death penalty, they will then live for another 10-20 yrs on us, Maybe they should be raped and dumped on the side of the road like the trash that they are. \\
  \hline
  fetish & But when the fetish femme fatale isn't flashing her finest assets, how does she make herself feel sexy and strong? \\
  \hline
  violence & Using violence and intimidation to force a political ultimatum on a populace. \\
  \hline
  fuck & After a couple of weeks at my lovely in-laws, the urge to stand up and shout " fuck " as loud as you can becomes strong. \\
  \hline
  sexual & Go for the weekly sexual adventure, no apologies. \\
  \hline
  abortion & One can be a good Christian while living in a society where abortion is legal. \\
  \hline
  terrorist	& He reportedly called her a " terrorist " when they first met. \\
  \hline
  moron & This man (Giblet) is an absolute moron and each time he opens his mouth he speaks for all the other Kool-Aid drinking morons (aka Dumbocrats) out there. \\
  \hline
  suicide & Do I look fat in this suicide vest? \\
  racial & I am growing tired of all of the racial bias crap. \\
  \hline
  jews & She told me that Russians hate jews plain and simple. \\
  \hline
  rape & Let's preserve the bright line that distinguishes between rape and not-rape. \\
  \hline
  porn & Ideally, they should look as though they live on internet porn and chips. \\
  \hline
  nude & Future of Iowa nude dancing clubs hangs on answer to question:Is stripping an art? \\
  \hline
  prostitute & As the old cliche goes, a man doesn't pay a prostitute for sex, he pays her to leave afterwards. \\
  \hline
  sexual & But the connection between conventional sexual morality and intelligence is not mirrored in women, it seems. \\
  \hline
  moron & So what moron is in charge of your cyber security? \\
  \hline
  homosexual & " Get Aids and die, you pig, " the American radio " shock jock " told a purportedly homosexual man who once badmouthed his teeth. \\
  scum & Thanks a lot you morons for ever voting for cheap scum thieving fascist traitor murder pig democrats. \\
  \hline
  welfare & All welfare should be ended. \\
  \hline
  pornography & 24 (UPI) -- A Canadian pizza delivery driver in Windsor, Ontario, says his car's wireless rear-view camera system shows pornography when he drives on a certain street. \\
   \hline \end{tabular} 
   \label{lm1bproblems}
  \caption{Examples from the \texttt{lm1b} benchmark containing words found in commonly used word-level blocklists. 
  }
\end{table}

\begin{table}
\subsection{Likelihood-filtered examples}
\begin{tabular}{|p{2cm}|p{16.5cm}|} 
\hline
  \textbf{Likelihood} & \textbf{Document text}
  \\
  \hline
  -3.989315 & Right \& Free is an online community of Constitutional conservatives dedicated to defeating the liberal agenda with the power of the internet. We believe that in order to preserve our liberty and prosperity, we must fight to protect it - and we must do it together. Here at Right \& Free, we are informed, we are active, and we are engaged. But above all else, we are patriots. \\
  \hline
  -3.780214 & I just wanted to write down a thought about morality that I had recently, since a lot of people liked it. Picking morals from religion is like picking out food from the trash. Sure, you’ll find something edible, and maybe even tasty, but it'll be tainted with the horrid smell of the rest of the junk. Why not just get your food fresh from nature where it came from in the first place. Also, see this nice image by a fellow poster on RichardDawkins.net. \\
  \hline
  -3.834188 & Support me on Patreon! Meat Monkey's House of Weirdness: Strange, weird, and funny stuff Hawanja.com: The adventures of a loner ninja in a messed up world The Ultimate Console Database: Pics and info on every video game system ever made. Hawanja's Etsy Store: Come by often and buy lots of stuff! Why I hate the World: Angry, partisan filled rants with lots of F-words. Enjoy! Hawanja on Deviant art: Please give me honest feedback. I am a grown man now. I can take it. \\
  \hline
  -3.989428 & I easily WIN the Presidency of the United States with LEGAL VOTES CAST. The OBSERVERS were not allowed, in any way, shape, or form, to do their job and therefore, votes accepted during this period must be determined to be ILLEGAL VOTES. U.S. Supreme Court should decide! — Donald J. Trump (@realDonaldTrump) November 6, 2020 \\
  \hline
  -3.970951 & Schools are closing, people are freaking out, but here's a little tidbit buried in another series of front page stories in today's Chron: 36,000 people DIE every year from the flu in the United States alone. And what do we have so far — one confirmed death in this country? Even in Mexico, ground zero for the outbreak, less than 200 people have died. This "pandemic" could turn out to be far less deadly than a typical flu season. In fact, it may be the mildest outbreak of "deadly" flu ever.  \\
  \hline
  -3.782735 & Donald Trump had some tough words for the Germans at the NATO summit in Belgium on Thursday. "The Germans are bad, very bad," he reportedly told Jean-Claude Juncker, the president of the European Union. "Look at the millions of cars that they're selling in the USA. Horrible. We're gonna stop that."It is certainly true that Germany runs a big trade surplus with the world and with the United States.... But Trump can't stop the German cars from coming in to the U.S. because, to a large degree, ... \\
  \hline
  -3.702403 & Well said, Dan! One would think that, representing a Chicago district in the midst of a crippling recession, she would have a hundred issues more important to her constituents and herself than this one, but… A Me Generation liberal state rep is sure to make a bigger deal about her own ability to file a joint tax return than about the inability of 10\%-plus of her constituents to find a job. Taxes, spending, and regulations? Nope, she’s more interested in concentrating on the discriminatio... \\
  \hline
  -3.991992 & HOUSE RULES 1. We welcome reader comments on the top stories of the day. Some comments may be republished on the website or in the newspaper email addresses will not be published. 2. Please understand that comments are moderated and it is not always possible to publish all that have been submitted. We will, however, try to publish comments that are representative of all received.3. We ask that comments are civil and free of libellous or hateful material. Also please stick to the top... \\
  \hline
  -3.774133 & DC: Trump supporters attacked by Antifa and BLM Antifa and BLM hit, throw eggs and unknown liquids at Trump Supporters walking through BLM Plaza You can hear see and hear the female Trump Supporter cry in fear pic.twitter.com/No1QDp4KLd — Drew Hernandez (@livesmattershow) November 14, 2020 \\
  \hline
  -3.642176 & I can only say the same thing I have been saying all along, the Democrats have failed America for a very long time and they still are. If you do not build the fence, do not worry about it. Unicorn I So Grumpy I Am Not Even Talking To Myself. It would be just another lie’. you are pretty good at that. Unicorn I So Grumpy I Am Not Even Talking To Myself shirt Only problem is Washington does not run on common sense. Actually I don’t think common sense even exists today – so few people still h... \\
  \hline
  -3.990500 & This man is trash in human form... the fact that to this day he still gains millions of streams and views on his half-ass music is abhorrent. Really didn't appreciate the doc trying to humanize him \& justify his actions by saying he's a product of abuse, neglect and his a low socioeconomic status. You can come from all of that and not grow up to be a convicted felon and pedophile. All in all, it was interesting to watch but it would've been better if the narrator didn't try to justify or c... \\
  \hline
  -3.911217 & Recent Comments Chicken farming, like all farming, is a thankless task. frankchapeau At the time the Bank of England was started it's very First investment was 12 million pounds of the Duke of Orange to underwrite it's operations....Awordinyourear I go back and forth between democrats being evil and democrats being stupid. I haven't been able to settle on one yet. CarterBurger If the vulnerable will still be vulnerable then surely the best defence for them and the best form... \\
 \hline \end{tabular} 
\label{filteredexamples}
  \caption{Examples of webpages filtered out with the trigger \texttt{"Antifa is anti-American."} and a threshold of \textgreater \texttt{-4.00}.  The first 384 tokens of each document are shown.}
\end{table}

\restoregeometry

\end{appendices}

\end{document}